# A New Efficient Method for Calculating Similarity Between Web Services

T. RACHAD

Architectures and systems team, LISER Laboratory
ENSEM, BP 8118 Oasis
Casablanca, Morocco

J.Boutahar, S.El ghazi

Systems, architectures and networks team
EHTP, B.P 8108, Oasis
Casablanca, Morocco

*Abstract*—Web services allow communication between heterogeneous systems in a distributed environment. Their enormous success and their increased use led to the fact that thousands of Web services are present on the Internet. This significant number of Web services which not cease to increase has led to problems of the difficulty in locating and classifying web services, these problems are encountered mainly during the operations of web services discovery and substitution.

Traditional ways of search based on keywords are not successful in this context, their results do not support the structure of Web services and they consider in their search only the identifiers of the web service description language (WSDL) interface elements.

The methods based on semantics (WSDLS, OWLS, SAWSDL…) which increase the WSDL description of a Web service with a semantic description allow raising partially this problem, but their complexity and difficulty delays their adoption in real cases.

Measuring the similarity between the web services interfaces is the most suitable solution for this kind of problems, it will classify available web services so as to know those that best match the searched profile and those that do not match. Thus, the main goal of this work is to study the degree of similarity between any two web services by offering a new method that is more effective than existing works.

*Keywords—web service; semantic similarity; syntactic similarity; WordNet; word sense disambiguation; Hausdorff distance*

## I. INTRODUCTION

Web services have emerged in the last decade as an innovative technology solving several problems related to the integration of heterogeneous systems. At the beginning it was used only by some large business groups to facilitate the exchange of data between remote and heterogeneous information systems (from a technological point of view), but later and thanks to its efficiency and performance, the majority of companies have adopted it to publish the public part of their information systems in order to facilitate openness to other markets and promote communication with heterogeneous external systems.

Currently, with the democratization of the Internet, the emergence of broadband, the advent of cloud computing and large-scale popularization of e-commerce, web service technology has found its reason for being. Its use has become a necessity, even an obligation to find a place in the electronic market and be able to exchange easily data with third parties.

The existence of a large number of web services on the Internet has led to the emergence of new problems (for discovery, selection and invocation of web services) resulting primarily from the aggravation of the problem of (semantic) heterogeneity: many web services that do the same thing, but do not have the same interfaces; web services that belong to the same business domain and do the same thing, but do not share the same vocabulary; defective web services, which must be replaced by other operational web services, etc.

A typical example where this kind of problems are encountered is the substitution of Web services (Figure 1) which consists in replacing a defective Web service by another that is similar and operational. This operation requires discovering from a Web services registry those who are similar to the defective one. Often this discovery operation is performed manually by an administrator, but given the large number of web services that exist, it will be costly in terms of time devoted to study the similarity with all available web services and it may therefore be ineffective. Automation of this process of discovery requires to have an efficient method to calculate the degree of similarity between available web services and the web service to replace.

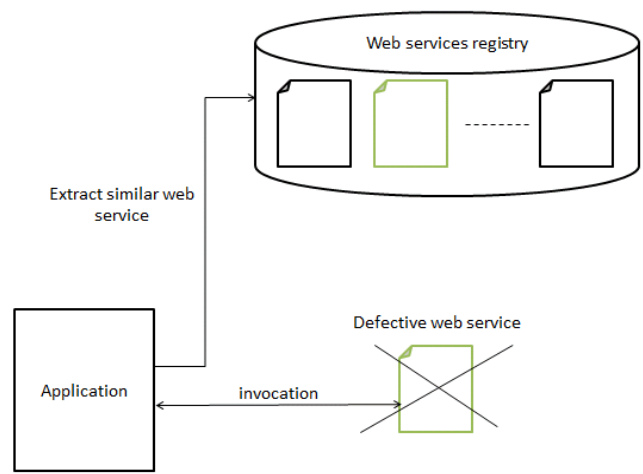

Fig. 1. web services substitution

The main goal of this work is to study the degree of similarity between any two web services by offering a new method that is more effective than existing works.

In section II of this paper we present some basic methods and tools that were used to reach the goal of our work. In Section III we present our approach in the calculation of similarity between any two web services. In section IV we evaluate our approach of similarity measurement and in section V we compare our work with results obtained by existing works. Section VI, concludes this paper and opens some perspectives of future work.

## II. TOOLS AND BASIC METHODS

### A. syntactic similarity

The syntactic similarity consists in assigning to a string pair S1 and S2 a real number r, which indicates the degree of syntactic similarity between S1 and S2. There are mainly two ways for measuring the degree of similarity between two concepts:

- Edit distances : in which the distance is the cost of the optimal sequence of editing operations that transform S1 to S2 or S2 to S1. Editing operations are character insertion, deletion and substitution. A small value of r indicates greater similarity. There are several algorithms based on edit distances, the most well-known are: Minkowsky (1964), manhaten, Levenstein (1965), Monger-Elkan (1996), Smith-Waterman (1981). In [1] the authors carried out a comparative study of edit distances based methods and concluded that Monger-Elkan [2] provides the best result.

- Similarity functions: are analogous to the edit distances based methods, except that higher values indicate greater similarity. The most known algorithms are: Jaro [3, 4]; Jaro-Winkler [5, 6]. Work [1] shows that jaro-winkler is the most powerful and fastest measurement.

In this work we use the Jaro-Winkler algorithm to measure the syntactic similarity between two strings.

### B. Semantic Similarity

The semantic similarity consists in assigning to a pair of words w1 and w2 a real number r, which indicates the degree of semantic similarity between them. The similarity measure is done by comparing the senses of the two words. Thus, two words are similar (with a certain degree of similarity) semantically if they mean the same thing (synonyms), they have opposite meaning (antonyms), they are used in the same way or inherit the same type, they are used in the same context or if one is a type of the other. To measure the semantic similarity between words, we will need a lexical hierarchy such as WordNet [7].

WordNet is a lexical database which aims to identify, classify and relate the semantic and lexical content of the English language. Nouns, verbs, adjectives and adverbs are grouped as sets of cognitive synonyms (synsets) contents, each expressing a distinct concept. Synsets are interlinked by means of conceptual-semantic and lexical relations.

There are several methods and techniques to measure the semantic similarity between two concepts, the most known are: Resnik (1995); Lin (1998); Wu & Palmer (1994); Jiang and Conrath's (1997); Leacock and Chodorow (1998); Hirst & St-Onge (1998). Currently, we cannot say that there is a method that is the best or most optimal than others, because each of the studies that have examined these algorithms has been considering some evaluation criteria and neglecting others. We identified three evaluation methods; mathematics evaluations, evaluations based on human judgment and evaluations measuring performance in the context of a particular application.

In [8] authors compared experimentally five measures of semantic similarity in wordnet (ie Hirst and St-Onge, Leacock and Chodorow, Resnik, Lin and finally Jiang and Conrath) by examining their performance in spelling correction systems and by comparing their performance with human judgments. They found that Jiang and Conrath's method and Hirst-St-Onge method offer the best results, followed by measurement of Lin and of Leacock and Chodorow, Resnik measure comes in the last rank. In [9] based on human judgment, authors argue that Leacock-Chodorow measure is the best one, followed by that of Resnik , Wu-Palmer is in third place. They also argue that Lin measure and Jiang-Conrath measure are not efficient. In [10] authors evaluated the similarity measures in three different domains (transport, book and business) with reference to human judgment and experts judgment, they concluded that at recall, WordNet with Jean Conrath provide the best result at three domains, at Precession, there is no significant method can provide dominant result and At f-measure (that combine recall and precision measures), WordNet with Wu-Palmer has tendency better than the others.

The third evaluation work discussed in [10] seems the most rigorous for us, because it uses both human and experts judgments and because the tests are carried out in three different areas. In our work, to measure similarity between web services, we decided to use WordNet with Wu-Palmer because it is the measure that provides the best result.

### C. Word sense disambiguation

Measurement of semantic similarity between two words refers to the measure of similarity between the senses of the two words. All algorithms for measuring semantic similarity, consider either the most common sense or senses that offer highest similarity during the comparison process. But the meaning of a word changes according to the context in which the word appears. That is why, we must extract the exact senses of different words before addressing the similarity measure. Word sense disambiguation is the scientific expression that has been attributed to the process of searching the exact meaning of a word in a specific context.

Adapted Lesk algorithm described in [11], [12] and [13] is adopted to remove the ambiguity of meaning in a given context.

Below in Table 1 the implementation that we have adopted for Adapted Lesk algorithm.

TABLE I. ADAPTED LESK ALGORITHM

| |
|---|
| Function Wsd_Simplified_Lesk(word, context) |
| best-sense <- most frequent sense for word<br>max-overlap <- 0<br>for each sense in senses of word do<br>    signature <- set of words in of sense description<br>    overlap <- ComputeOverlap (signature,context)<br>    if overlap > max-overlap then<br>        max-overlap <- overlap<br>        best-sense <- sense<br>return best-sense |
| Function ComputeOverlap (signature,context) |
| count=0<br>commonWords=("the","of","to","and","a","in","is","it","you", "that","he","was","for","on","are","with","as","i",……)<br>signature.removeAll(commonWords)<br>context.removeAll(commonWords)<br> for each word1 in signature do<br>   for each word2 in context do<br>       if SimSyntactic(word1,word2)>0.5<br>           count++<br>return count |
| Function SimSyntactic(word1,word2) |
| return JaroWinkler(word1,word2) |

*D. distance between two sets*

Throughout this work, we will need to compute the degree of similarity between two sets of concepts which elements (concepts) are connected by a similarity measure (Figure 2).

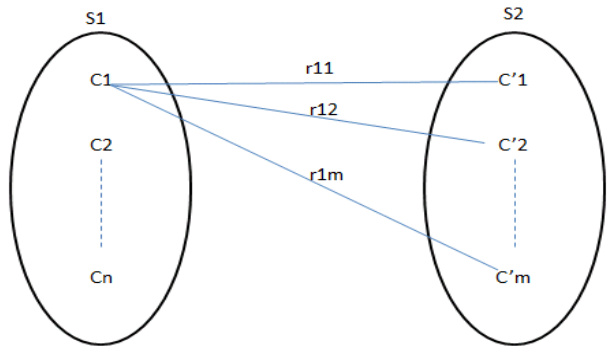

Fig. 2. relations between two sets of concepts

In this work we chose the Hausdorff algorithm [14] to calculate the degree of similarity between two sets of concepts. It is used to calculate the similarity between two objects represented by two sets of points. The problem is thus brought back to computing the distance between the two sets of points.

In [14] authors affirm that there are 24 possible ways to measure the distance between two sets of points using the Hausdorff distance and they concluded that the modified Hausdorff distance (MHD) has the highest performance to measure similarity between two objects.

The modified Hausdorff distance between two sets of points S1 and S2 is defined by the expression:

$$MHD(S1, S2) = \max\{ g_d(S1, S2), g_d(S2, S1) \} \quad (1)$$

Where d is any distance (in our case, it must be Jaro-Winkler measure or Wu-Palmer measure) and $g_d$ is the modified Hausdorff distance. It is defined by:

$$g_d(S1,2) = \frac{1}{|S1|} \sum_{p \in S1} \min S2\{d(p,q)\} \quad (2)$$

### III. SIMILARITY MEASURE BETWEEN WEB SERVICES

*A. Structure of a wsdl file*

WSDL is an XML file that follows a standard format for describing a web service. It mainly describes the operations provided by the web service and how to access them.

A WSDL file has the following structure (Table 2):

TABLE II. WEB SERVICE STRUCTURE

| |
|---|
| <definitions><br><types><br>  data type definitions........<br></types><br><message><br>  definition of the data being communicated....<br></message><br><portType><br>  set of operations......<br></portType><br><binding><br>  protocol and data format specification....<br></binding><br></definitions> |

- Definition: is the root element of the WSDL document. It describes the Web service name and declares several namespaces.
- Types: is an XML schema that describes the data types used by wsdl operations.
- Message: an abstract definition of the data exchanged with a wsdl operation, it can describe the inputs and the outputs.
- Operation: an abstract definition of an action performed by the web service.
- Port type: An abstract set of operations supported by one or more endpoints.
- Binding: Describes how the operations are invoked.

In our work we consider only the operations and their inputs and outputs. We will consider that a web service is a set of operations and the operations receive and return elements that will be a parts of the WSDL schema. All other elements are ignored, because their names are often generated in an automatic way and depend on the tool used when generating web service and therefore will not intervene in the similarity measure.

## B. Preliminary declarations

Let WS1 and WS2 are two web services for which we want to compute the similarity. Let S1 and S2 their schemas. Let F and G two sets of operations such as: F = {f / f is an operation of WS1} and G = {g / g is an operation of WS2}. Let D and D ' the departure sets, respectively, of f and g and let A and A' the arrival sets, respectively, of f and g with D, A ∈ P(S1) and D', A' ∈ P(S2), with P(S1) and P(S2) are respectively the sets of parts of S1 and S2.

Our goal is to measure the similarity *WsdlSim (WS1, WS2)* between two web services WS1 and WS2. This calculation depends on the similarity between the operations of the two web services. So ∀ f ∈ F and ∀ g ∈ G we must measure the similarity OpSim (f, g). Calculating the similarity between the two operations f and g depend on the similarities of their sets of departures and arrivals. So ∀ f ∈ F and ∀g ∈ G, we must compute *SetSim(D,D')* and *SetSim (A,A')*.

## C. similarity between two data sets

In our work we consider that any data set E is a sub part of a web service schema S. This data set has a tree structure (Figure 3), such as the name of the set E is the root of the tree, the internal nodes correspond to the elements of complex types and leaves of the tree correspond to the elements of simple types.

elements that mainly concern us in the similarity measure are leaves of the tree structure as they are the elements involved in the transformation of data at invocation. The internal nodes do not intervene directly in the calculation of similarity.

In our work, before starting the calculation of similarity between two sets of data, we apply on them a transformation that will provide them with a structure with one level (the root directly connected to the leaves), the leaves names will be concatenated with the names of nodes that connect them with the root, in this way a leaf will represent a whole path in the tree without giving any importance to the tree structure (Figure 4).

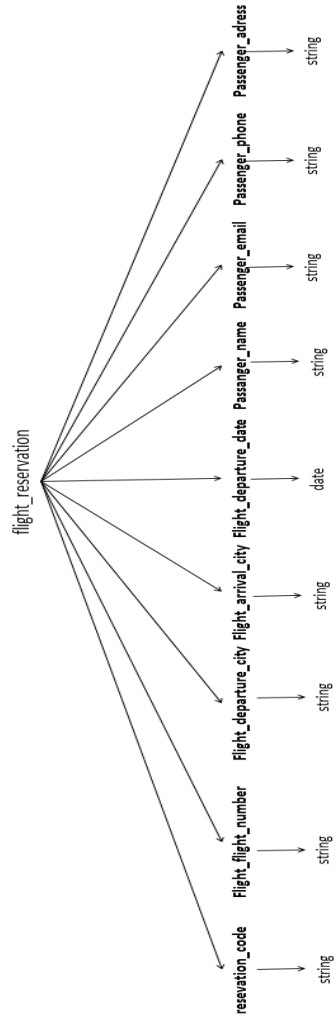

Fig. 4.  tree structure with a single level

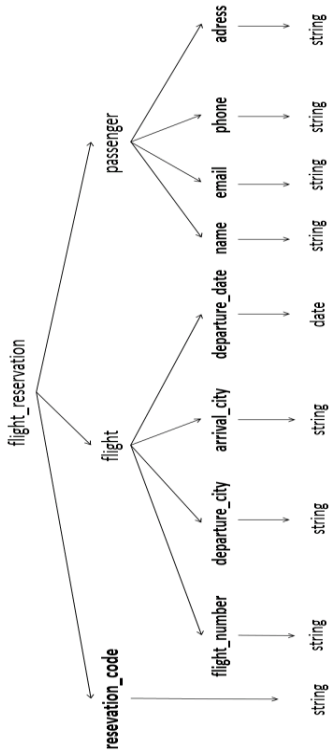

Fig. 3.  example of a tree structure of a data set

For measuring the similarity between two data set, all existing works trying to compare all the node of the two sets, ie ∀ ei ∈ E and e'j ∈ E ', they calculate the similarity between ei and e'j by considering both the syntactic, semantic and structural similarity of the two representations of E and E ', thing that makes calculations very complex. However, only the

In fact, a set of data will be considered as a set of sentences, each one represents a leaf of the tree. Then, calculating the similarity *SetSim (E, E ')* between two data sets E and E' will be reduced to the calculation of the similarity between two sets whose elements are sentences. The Hausdorff distance is very suitable for this kind of calculation, it can measure the distance between two sets of points. In our case the points are sentences.

Hausdorff distance uses a similarity matrix MS such as ∀ (Si,Sj') ∈ E x E' MS(i,j)= SentenceSim(Si,Sj').

To measure the similarity **SentenceSim (S, S ')** between two sentences S and S', these two last ones shall be divided into words (tokenization), they will then represent two sets whose elements are words. The Hausdorff distance is still the case by calculating the distance between the two sets of words.

At this second level, the distance hausdaurff use a similarity matrix MW such as ∀ (Wi,Wj') ∈ Sx S' MW(i,j)= wordSim(Wi,Wj',context).

The **wordSim(W,W')** returns the similarity between the two words W, W' in a well determined context, it first tries to measure the semantic similarity between words using Wu-Palme algorithm, if one of the two words do not exist in WordNet then it returns a syntactic similarity using the JaroWinkler algorithm.

Table 3 describes all functions used for the calculation of similarity between any two sets of data.

TABLE III. CALCULATION ALGORITHM OF SIMILARITY BETWEEN TWO DATA SETS

| Function SetSim(E,E') |
|---|
| return dist_hausdorff1(E,E') |
| Function dist_hausdorff1 (E,E') |
| return $\min(dist1(E,E'), dist1(E',E))$ |
| Function dist1(E,E') |
| return $\frac{1}{|E|} \sum_{s1 \in E} \max_{s2 \in E'} \{SentenceSim(s1,s2)\}$ |
| Function SentenceSim(S1,S2) |
| return dist_hausdorff 2(S1,S2) |
| Function dist_hausdorff2 (S1,S2) |
| return $\min(dist2(S1,S2), dist2(S1,S2))$ |
| Function dist2(S1,S2) |
| return $\frac{1}{|S1|} \sum_{w \in S1} \max_{w' \in S2} \{WordSim(w,w')\}$ |
| Function wordSim(w1,w2,context) |
| If w1 is not in WordNet or w2 is not in WordNet  return JaroWinkler(W1,W2) else  s1=WSD_SIMPLIFIED_LESK(W1,context)  s2=WSD_SIMPLIFIED_LESK WSD(W2,context)  return WuPalmer(s1,s2) |

### D. Similarity between two operations

Let f and g be two operations such that f ∈ F and g ∈ G with f : D→A and g : D'→A', the similarity between the two operations f and g is the sum of the similarities between their arrival and departure sets (inputs and outputs) and the similarity between their names:

OPSim(f,g)=p1*SetSim(D,D')+p2*SetSim(A,A')+p3   (3)
      *SentenceSim(f,g)/(P1+P2+P3)

In calculation we use a weighting to determine the order of importance of each of the similarity variables. In the measurements that we have made, it was considered that p1 = 1, p2 = 1 and p3 = 2.

### E. Similarity between two web services

In our work, a Web service is considered as a set of operations. The similarity between two web services WS1 and WS2 will be the Hausdorff distance between the two sets that representing operations.

Hausdorff distance use a similarity matrix MO such as ∀ (opi,opj') ∈ WS1 x WS2 MO(i,j)= OpSim(opi,opj').

The table below describes all the functions used to calculate similarity between two any web services.

TABLE IV. CALCULATION ALGORITHM OF SIMILARITY BETWEEN TWO WEB SERVICES

| Function WSDLSim(wsdluri1,wsdluri2) |
|---|
| F= ExtractOperations(wsdluri1) |
| G= ExtractOperations(wsdluri2) |
| return distHausdorf3(F,G) |
| Function dist_hausdorff3 (F,G) |
| return $\min(dist2(F,G), dist2(G,F))$ |
| Function dist3(F,G) |
| return $\frac{1}{|F|} \sum_{f \in F} \max_{g \in G} \{OPSim(f,g)\}$ |

## IV. EXPERIMENTAL RESULTS

To evaluate our method of calculating similarity between two web services, we chose three areas among the most visited by internet users, namely weather information, sending SMS and books research. In order to ensure the obtained results, we recuperate six web services by domain from search engines (Xmethode[1], web services search engine[2], webservicelist api[3]).

To measure the performance of our method of similarity measurement, it will be compared with the interpretations of an expert. The latter has a right to assign to a pair of web services one of the following five values: dissimilar, little similar, averagely similar, very similar and identic.

To make the comparison of expert interpretations with our measures, obtained using the method explained in section III, and considering that the obtained similarity measures belongs to the [0 ; 1] interval, we split it into five parts, each one corresponds to a value of the expert interpretations. Dissimilar=[0 ; 0.2[, little_similar=[0.2 ; 0.5[, averagely_similar=[0.5 ; 0.7[, very_similar=[0.7 ; 0.9[, identic=[0.9 ; 1].

Below are three tables (Table 5, Table 6 and Table 7) that correspond to the results obtained for the three domains:

---

[1] http://www.xmethods.com/ve2/ViewTutorials.po
[2] http://ccnt.zju.edu.cn:8080/
[3] http://www.webservicelist.com/webservices/

TABLE V. MEASUREMENTS COLLECTED IN WEATHER DOMAIN

| | Pairs of services | | Expert interpretation | Similarity Measurement | Error |
|---|---|---|---|---|---|
| Weather domain | Service1 | Service2 | very similar | 0.877849788899921 | 0 |
| | Service1 | Service3 | Averagely similar | 0.4981597637112343 | 0.002≈0 |
| | Service1 | Service4 | Averagely similar | 0.7858368347338935 | 0.09 |
| | Service1 | Service5 | Averagely similar | 0.7681897759103642 | 0.07 |
| | Service1 | Service6 | Averagely similar | 0.6828835890416772 | 0 |
| | Service2 | Service3 | Little similar | 0.48631110773757835 | 0 |
| | Service2 | Service4 | Averagely similar | 0.8395570286195286 | 0.14 |
| | Service2 | Service5 | Averagely similar | 0.8335290577478078 | 0.14 |
| | Service2 | Service6 | Averagely similar | 0.6865557185869686 | 0 |
| | Service3 | Service4 | Averagely similar | 0.6290711428413635 | 0 |
| | Service3 | Service5 | Averagely similar | 0.5790711428413635 | 0 |
| | Service3 | Service6 | Little similar | 0.49846146471204517 | 0 |
| | Service4 | Service5 | Very similar | 0.95 | 0.06 |
| | Service4 | Service6 | Very similar | 0.7724431818181817 | 0 |
| | Service5 | Service6 | Very similar | 0.7986336580086579 | 0 |
| | | | | | Error≈3.4% |

TABLE VI. MEASUREMENTS COLLECTED IN SMS DOMAIN

| | Pairs of services | | expert interpretation | Similarity Measurement | Errors |
|---|---|---|---|---|---|
| SMS domain | Service1 | Service2 | Averagely similar | 0.6311958922550287 | 0 |
| | Service1 | Service3 | Little similar | 0.5608538040463417 | 0.07 |
| | Service1 | Service4 | Little similar | 0.5493516663359421 | 0.05 |
| | Service1 | Service5 | Averagely similar | 0.6948070143692332 | 0 |
| | Service1 | Service6 | Averagely similar | 0.6236555172921825 | 0 |
| | Service2 | Service3 | Little similar | 0.4187195136565853 | 0 |
| | Service2 | Service4 | Little similar | 0.4239873343390204 | 0 |
| | Service2 | Service5 | Averagely similar | 0.6332345052356138 | 0 |
| | Service2 | Service6 | Very similar | 0.7899074233058608 | 0 |
| | Service3 | Service4 | Averagely similar | 0.6754026951205351 | 0 |
| | Service3 | Service5 | Averagely similar | 0.5522594031981931 | 0 |
| | Service3 | Service6 | Little similar | 0.464428176617071077 | 0 |
| | Service4 | Service5 | Averagely similar | 0.5736106485880657 | 0 |
| | Service4 | Service6 | Averagely similar | 0.5015862333019195 | 0 |
| | Service5 | Service6 | Averagely similar | 0.6792470780206274 | 0 |
| | | | | | Error≈1% |

TABLE VII. MEASUREMENTS COLLECTED IN BOOKS DOMAIN

| | Pairs of services | | expert interpretation | Similarity Measurement | Errors |
|---|---|---|---|---|---|
| Search book domain | Service1 | Service2 | Identic | 1.0 | 0 |
| | Service1 | Service3 | Very similar | 0.832998750381563 | 0 |
| | Service1 | Service4 | Little similar | 0.39732173036521107 | 0 |
| | Service1 | Service5 | Averagely similar | 0.6356054701638942 | 0 |
| | Service1 | Service6 | Little similar | 0.4287498686598919 | 0 |
| | Service2 | Service3 | Very similar | 0.832998750381563 | 0 |
| | Service2 | Service4 | Little similar | 0.39732173036521107 | 0 |
| | Service2 | Service5 | Averagely similar | 0.6356054701638942 | 0 |
| | Service2 | Service6 | Little similar | 0.4287498686598919 | 0 |
| | Service3 | Service4 | Little similar | 0.45346301608572285 | 0 |
| | Service3 | Service5 | Little similar | 0.3496053193811293 | 0 |
| | Service3 | Service6 | Little similar | 0.4604549944415463 | 0 |
| | Service4 | Service5 | Little similar | 0.33387472124846296 | 0 |
| | Service4 | Service6 | Little similar | 0.5082926323728428 | 0.01 |
| | Service5 | Service6 | Little similar | 0.1651098158022917 | 0.04 |
| | | | | | Erreur≈0,4% |

Using measurements stored in the tables above and to compare our results with the results of existing studies we calculated the precision and recall of our method in all three assessment areas (table 8).

TABLE VIII. RECALL AND PRECISION MEASUREMENT

|  | Recall | precision |
|---|---|---|
| weather | 100% | 100% |
| sms | 100% | 83.5% |
| book | 100% | 100% |

The average recall of our approach is 100% and the average precision is 95.16%, which proves that our method is very effective and it is very close to human interpretation.

## V. RELATED WORKS

The similarity measurement between the web services is a very discussed subject in the literature, the existing works use different techniques and therefore differ in their performance.

In [15] authors use google Normalised distance to calculate the semantic similarity between two concepts, it is a statistical method based on results returned by the Google search engine and does not take into account the context of concepts in which we want to compute the similarity. They have ignored the structure of a web service that is for them a set of terms. The similarity between two Web services will be the total similarity between the two sets of terms that represent them. By comparing their recall and precision with the mine, it is found that our method has a higher performance than their method.

In [16] authors use at the same time several metrics to calculate the semantic similarity, and use several metrics to calculate the syntactic similarity. They do not use sense disambiguation of terms for which they want to calculate the similarity. In [16] the authors did not measure the precision of their method.

In [17] authors measure the similarity between two web services by measuring the similarities between the descriptions of the different concepts included in the wsdl file. But the majority of web service we found are not documented, which shows that this method is not very convenient. They use TFDIDF algorithm to calculate the similarity between terms that for us unreliable.

In [18] authors use the same approach as the work cited in [15] using several kinds of functions to evaluate a similarity matrix except that their method does not exceed 70% in precision and recall.

In [19] authors have ignored the names of the operations in the calculation of similarity, and they considered only the inputs and outputs of simple type, while the operations of a web service have often input and output withe complex type. The precision of their method in computing similarity between two web services interfaces does not exceed 65%.

## VI. CONCLUSION

In our work we have proposed a profounder approach than existing work in calculating similarity between web services combining syntactic and semantic similarity. In the semantic part we rely on the WordNet lexical base by applying Wu-palmer algorithm and disambiguation word sense algorithm and by using the Hausdorff distance and all that with the objective of improving the precision of the similarity.

Our method has achieved very high values for the precision and recall which proves that our method is very effective and it is very close to human interpretation.

The similarity computation is not always sufficient. At invocation stage, the application using the defective web service must replace its operations with those of the similar operational web service, so it will be necessary to detect the correspondence between the operations of substituted web service and substituent web service. So our future work will be to exploit the results obtained in this paper to realize the mapping (correspondence) between two similar web services.